\documentclass{article}
\usepackage{ijcai20}

\usepackage{times}

\usepackage{soul}
\usepackage{url}
\usepackage[hidelinks]{hyperref}
\usepackage[utf8]{inputenc}
\usepackage[small]{caption}
\usepackage{graphicx}
\usepackage{amsmath}
\usepackage{booktabs}
\usepackage{multirow}
\usepackage{subfig}
\usepackage{rotating}

\makeatletter
\newif\if@restonecol
\makeatother

\usepackage[linesnumbered,ruled,vlined]{algorithm2e}
\usepackage{algpseudocode}

\title{CDIMC-net: Cognitive Deep Incomplete Multi-view Clustering Network}

\author{
Jie Wen$^{1,2{\dagger}}$\and
Zheng Zhang$^{1,2,3{\dagger}}$\and
Yong Xu$^{1,2,3}$\footnote{corresponding author, `$\dagger$' indicates co-first authors.}\and
Bob Zhang$^4$\and
Lunke Fei$^5$\And
Guo-Sen Xie$^6$\\
\affiliations
$^1$Bio-Computing Research Center, Harbin Institute of Technology, Shenzhen, Shenzhen, China\\
$^2$Shenzhen Key Laboratory of Visual Object Detection and Recognition, Shenzhen, China\\
$^3$Pengcheng Laboratory, Shenzhen, China\\
$^4$Department of Computer and Information Science, University of Macau, Taipa, Macau, China\\
$^5$School of Computer Science and Technology, Guangdong University of Technology, Guangzhou, China\\
$^6$Inception Institute of Artificial Intelligence, Abu Dhabi, UAE\\
\emails
jiewen\_pr@126.com,
darrenzz219@gmail.com,
yongxu@ymail.com,
bobzhang@um.edu.mo,
flksxm@126.com,
gsxiehm@gmail.com
}
\begin{document}

\maketitle

\begin{abstract}
In recent years, incomplete multi-view clustering, which studies the challenging multi-view clustering problem on missing views, has received growing research interests. Although a series of methods have been proposed to address this issue, the following problems still exist: 1) Almost all of the existing methods are based on shallow models, which is difficult to obtain discriminative common representations. 2) These methods are generally sensitive to noise or outliers since the negative samples are treated equally as the important samples. In this paper, we propose a novel incomplete multi-view clustering network, called Cognitive Deep Incomplete Multi-view Clustering Network (CDIMC-net), to address these issues. Specifically, it captures the high-level features and local structure of each view by incorporating the view-specific deep encoders and graph embedding strategy into a framework. Moreover, based on the human cognition, \emph{i.e.}, learning from easy to hard, it introduces a self-paced strategy to select the most confident samples for model training, which can reduce the negative influence of outliers. Experimental results on several incomplete datasets show that CDIMC-net outperforms the state-of-the-art incomplete multi-view clustering methods.
\end{abstract}

\section{Introduction}
Multi-view clustering is a well-known research topic in fields of machine learning \cite{chao2017survey,zhang2018binary}. Generally speaking, almost all of the previous researches on multi-view clustering are based on the assumption that all views of samples are available and strictly aligned. However, in practical applications, more and more collected multi-view data are incomplete where some views are unavailable. For example, many volunteers only have one or two kinds of examination results of magnetic resonance imaging, positron emission tomography, and cerebrospinal fluid for Alzheimer’s disease diagnosing \cite{xiang2013multi}. Multi-view data with missing views are called \emph{incomplete multi-view data} and clustering on such data is the so-called \emph{incomplete multi-view clustering} (IMC) \cite{wen2019unified,hu2018doubly}. Obviously, conventional methods fail to handle these incomplete multi-view data. In addition, owing to the missing views, it is difficult to explore the complementary and consistent information from the incomplete multi-view data, which makes IMC a very challenging task.

For IMC, Trivedi et al. \cite{trivedi2010multiview} proposed a kernel canonical correlation analysis (KCCA) based method, which recovers the kernel matrix of the incomplete view according to that of the complete view. However, it can only handle the two-view data and requires that one view is complete. To address this issue, Li et al. proposed the partial multi-view clustering (PMVC) based on matrix factorization, where the paired views are decomposed into the same representation \cite{li2014partial}. Then various extensions of PMVC, such as graph regularized PMVC (GPMVC) \cite{rai2016partial} and incomplete multi-modal grouping (IMG) \cite{zhao2016incomplete} have been proposed, which mainly incorporate the graph embedding technique to enhance the separability of the common representation. In \cite{zhao2018incomplete}, the graph embedding and deep feature extraction techniques are simultaneously integrated into PMVC to preserve the local structure and capture the high-level features. Partial multi-view clustering via consistent GAN (PMVC\_GAN) unifies the Autoencoder and cycle generative adversarial network (GAN) into a novel IMC framework, which can infer the missing views via GAN and in turn promotes the common representation learning \cite{wang2018partial}. However, the above methods are not applicable to arbitrary incomplete cases where some incomplete samples have more than one views or no samples have complete views \cite{wen2018incompleteTCYB}. To address the issue, weighted matrix factorization technique is introduced for IMC, where the representative works are multiple incomplete views clustering (MIC) \cite{shao2015multiple}, doubly aligned IMC (DAIMC) \cite{hu2018doubly}, one-pass IMC (OPIMC) \cite{hu2019onepass}, and online multi-view clustering (OMVC) \cite{shao2016online}. Besides these methods, many graph learning based methods~\cite{wang2019spectral,wen2018incompleteTCYB} and multiple kernels based methods (such as incomplete multiple kernel k-means with mutual kernel completion (MKKM-IK-MKC) \cite{liu2019multiple}) have also been proposed to handle the arbitrary IMC cases.

Although the aforementioned methods provide some schemes to address the IMC problem, these methods still suffer from the following issues: 1) most of the previous methods exploit the shallow models to obtain the common representation, which cannot capture the high-level features from the complex multi-view data. Although some deep network based methods like PMVC\_GAN can capture the high-level features, it is inflexible to handle all kinds of incomplete cases. 2) None of them considers the negative effect of marginal samples since these methods treat all samples equally. This is unreasonable because those marginal samples are generally far away from the cluster centers and can be viewed as outliers, which are harmful to model training \cite{guo2019adaptive}. In this paper, we propose a novel cognitive based deep incomplete multi-view clustering network, referred to as CDIMC-net, to address the above issues. CDIMC-net integrates several view-specific deep encoders, a self-paced kmeans clustering layer, and multiple graph constraints into a unified network for arbitrary IMC. Representatively, the main contributions of our work are illustrated as follows:

1) We propose a novel and flexible deep clustering network for arbitrary IMC cases, which incorporates the graph embedding to promote the network training.

2) This is the first work that introduces the human cognitive based learning into IMC. Compared with the existing works, CDIMC-net can adaptively reduce the negative influence of the marginal samples, and thus is more robust to outliers.
\section{Kmeans Clustering}
Kmeans is one of the most famous clustering algorithms. For any data $X = \left[ {{x_1}, \ldots ,{x_n}} \right] \in {R^{m \times n}}$ with $n$ samples and $m$ features, kmeans seeks to find $k$ optimal cluster centers $U = \left[ {{u_1}, \ldots ,{u_k}} \right] \in {R^{m \times k}}$ and cluster indicator $S \in {\left\{ {0,1} \right\}^{k \times n}}$ by solving the following problem \cite{nie2019k}:
\begin{equation}\label{eq_2}
\setlength{\abovedisplayskip}{3pt}
\setlength{\belowdisplayskip}{3pt}
\mathop {\min }\limits_{U,S} \left\| {X - US} \right\|_F^2{\kern 1pt} {\kern 1pt} {\kern 1pt} {\kern 1pt} {\kern 1pt} s.t.{\kern 1pt} {\kern 1pt} {\kern 1pt} {\kern 1pt} S \in {\left\{ {0,1} \right\}^{k \times n}},S^{T}\emph{1}  = \emph{1}
\end{equation}
where ${S_{j,i}} = 1$ denotes that the corresponding $i$-th sample ${x_i}$ is partitioned into the $j$-th cluster. $\emph{1}$ is a vector with all elements as 1.

\section{The Proposed Method}
\subsection{Problem Statement}
For the given incomplete multi-view data with $l$ views, we use ${X^{\left( v \right)}} = \left[ {x_1^{\left( v \right)}, \ldots ,x_n^{\left( v \right)}} \right] \in {R^{{m_v} \times n}}$ to represent the instance set of the $v$-th view, where ${m_v}$ is the feature dimension, $n$ denotes the number of samples, and elements of the missing instances are denoted as `NaN' (\emph{i.e.}, not a number). The view available and missing information is recorded in a diagonal matrix ${W^{\left( v \right)}}$ for the $v$-th view, where $W_{i,i}^{\left( v \right)} = 1$ if the $i$-th instance is available in the $v$-th view, otherwise $W_{i,i}^{\left( v \right)} = 0$. The goal of IMC is to group these $n$ samples into $k$ clusters.

\subsection{CDIMC-net}
As shown in Fig.1, CDIMC-net groups the incomplete multi-view data via two phases: pre-training and fine-tuning, where an Autoencoder based pre-training phase is used to initialize the network parameters and the fine-tuning phase aims at obtaining the cluster-friendly representations while producing the cluster indicators for all input samples.

\begin{figure*}[th]
\begin{center}
\includegraphics[width=0.65\linewidth]{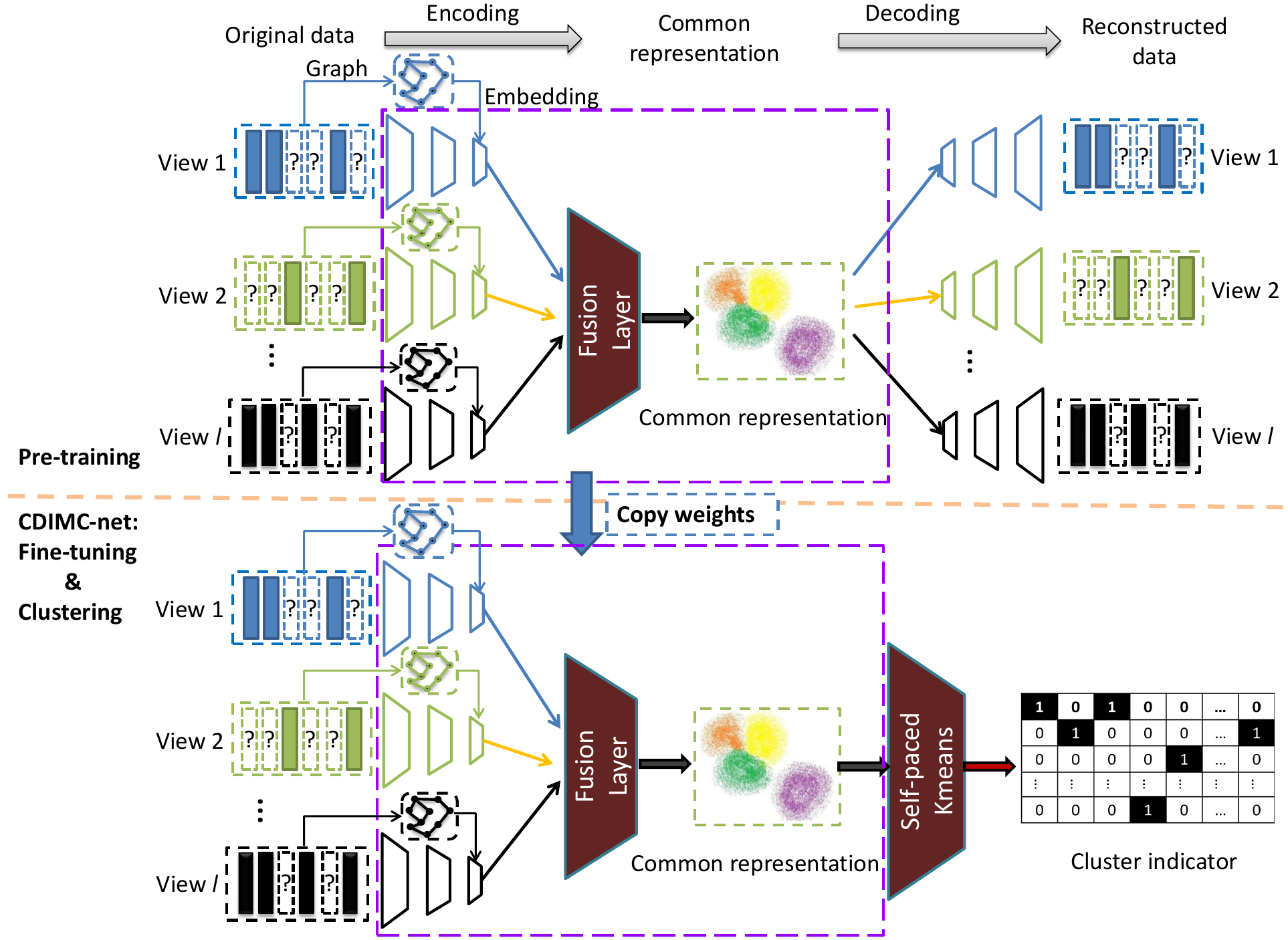}
\end{center}
\caption{The proposed CDIMC-net for incomplete multi-view clustering.}
\label{fig:fig1}
\end{figure*}

\subsubsection{Pre-training Network}
In our work, based on the conventional under-complete Autoencoder~\cite{guo2017improved,xie2016unsupervised}, we develop a graph regularized incomplete multi-view Autoencoder for incomplete multi-view cases, where the graph embedding technique is introduced to preserve the local structure of data and a weighted fusion layer is introduced to eliminate the negative influence of missing views. Specifically, the proposed incomplete multi-view Autoencoder is composed of the following three components.

\textbf{View-specific encoders and decoders}: As the basic Autoencoder, encoder network captures the most salient features from the high-dimensional data and the decoder network aims at recovering the data from the encoded features~\cite{guo2017improved}. Considering that different views may have different dimensions, information, structures, and appearances, we design several view-specific encoders {\small{$\left\{ {f_{EC}^{\left( v \right)}} \right\}_{v = 1}^l$}} and corresponding decoders {\small{$\left\{ {f_{DC}^{\left( v \right)}} \right\}_{v = 1}^l$}} for different views.

\textbf{Fusion layer}: As mentioned in many references, all views share the common semantic information for the same sample, such as the common representation or cluster label~\cite{zhao2018incomplete,hu2019onepass}. Inspired by this, CDIMC-net seeks to obtain the common representation shared by all views for clustering. Specifically, a simple weighted fusion layer is introduced for such goal as follows:
\begin{equation}\label{eq_4}
h_i^* = {{\sum\limits_{v = 1}^l {W_{i,i}^{\left( v \right)}h_i^{\left( v \right)}} } \mathord{\left/
 {\vphantom {{\sum\limits_{v = 1}^l {W_{i,i}^{\left( v \right)}h_i^{\left( v \right)}} } {\sum\limits_{v = 1}^l {W_{i,i}^{\left( v \right)}} }}} \right.
 \kern-\nulldelimiterspace} {\sum\limits_{v = 1}^l {W_{i,i}^{\left( v \right)}} }}
\end{equation}
where $h_i^{(v)}$ is the output of instance $x_i^{( v)}$ at the $v$-th encoder $f_{EC}^{( v)}$. $h_i^*$ is the common representation for the $i$-th sample.

As shown in Fig.1, the fusion layer is placed between the encoding network and the decoding network. Introducing the weighted fusion layer can solve the issue of incomplete learning by reducing the negative influence of missing views.

\textbf{Graph embedding}: In fields of subspace learning, a recognized manifold assumption is that if two data points ${x_i}$ and ${x_j}$ are close to each other, then their corresponding low dimensional representations should also be close in the latent subspace~\cite{cai2008non,kang2017clustering,wen2018low}. To preserve such neighbor relationships, the following graph embedding constraint is considered:
\begin{equation}\label{eq_5}
\setlength{\abovedisplayskip}{3pt}
\setlength{\belowdisplayskip}{3pt}
\min \frac{1}{{2nl}}\sum\limits_{v = 1}^l {\sum\limits_{i = 1}^n {\sum\limits_{j = 1}^n {\left\| {h_i^{\left( v \right)} - h_j^{\left( v \right)}} \right\|_2^2N_{i,j}^{\left( v \right)}} } }
\end{equation}
where $N^{\left( v \right)} \in {R^{n \times n}}$ denotes the nearest neighbor graph constructed from instance set ${X^{\left( v \right)}}$ as follows:
\begin{equation}\label{eq_6}\small
N_{i,j}^{( v )} = \left\{ {\begin{array}{*{20}{c}}
{1,}&\begin{array}{l}
if{\kern 1pt} {\kern 1pt} ( {x_i^{( v )} \ne NaN,x_j^{( v )} \ne NaN} )\& \\
{\kern 1pt} ( {x_i^{( v )} \in \psi ( {x_j^{( v )}} ){\kern 1pt} {\kern 1pt} or{\kern 1pt} {\kern 1pt} x_j^{( v )} \in \psi ( {x_i^{( v )}} )} ){\kern 1pt} {\kern 1pt}
\end{array}\\
{0,}&{otherwise}
\end{array}} \right.
\end{equation}
where $\psi ( {x_i^{( v)}})$ denotes the nearest instance set to $x_i^{( v)}$.

\textbf{Loss function of pre-training}: Combining the graph embedding loss and Autoencoder loss, the overall loss function of the pre-training network is designed as:
\begin{align}\label{eq_7}
\mathop {\min }\limits_{{\Omega _1},{\Omega _2}} &\sum\limits_{v = 1}^l {\frac{1}{{{m_v}n}}} \left\| {\left( {{X^{\left( v \right)}} - {{\bar X}^{\left( v \right)}}} \right){W^{\left( v \right)}}} \right\|_F^2 \nonumber\\
 &+ \alpha \frac{1}{{2nl}}\sum\limits_{v = 1}^l {\sum\limits_{i = 1}^n {\sum\limits_{j = 1}^n {\left\| {h_i^{\left( v \right)} - h_j^{\left( v \right)}} \right\|_2^2N_{i,j}^{\left( v \right)}} } }
\end{align}
where {\small{${\bar X^{\left( v \right)}} = f_{DC}^{\left( v \right)}\left( {f_{EC}^{\left( v \right)}\left( {{X^{\left( v \right)}}} \right)} \right)$}} denotes the reconstructed data of the $v$-th view, ${\Omega _1}$ and ${\Omega _2}$ denote the parameters of encoder network and decoder network, respectively. $\alpha$ is a positive hyper-parameter.
\subsubsection{Fine-tuning and Clustering}
By optimizing problem (\ref{eq_7}), we can obtain a more compact common representation ${H^*}$ for the incomplete multi-view data and the initialized network parameters. However, it cannot guarantee the obtained common representation to be cluster-friendly. In many previous deep clustering works, kmeans is widely considered~\cite{yang2017towards,guo2018deep}. However, as can be seen from objective function (\ref{eq_2}), the conventional kmeans treats all samples including the cluster-oriented samples and marginal samples (outliers) equally, which makes the trained clustering network be sensitive to outliers. A good approach to address this issue is to select the most cluster-oriented samples for model fine-tuning~\cite{guo2019adaptive}. To this end, we introduce a self-paced kmeans as the clustering layer for fine-tuning.

\textbf{Self-paced kmeans}: The loss function of the self-paced kmeans is denoted as follows~\cite{guo2019adaptive}:
\begin{equation}\label{eq_8}
\setlength{\abovedisplayskip}{3pt}
\setlength{\belowdisplayskip}{3pt}
\begin{array}{l}
\mathop {\min }\limits_{S,v,\lambda } \frac{1}{nk}\sum\limits_{i = 1}^n {\left( {{r_i}\left\| {h_i^* - U{S_{:,i}}} \right\|_2^2 - \lambda {r_i}} \right)} \\
s.t.{\kern 1pt} {\kern 1pt} {r_i} \in \left\{ {0,1} \right\},S \in {\left\{ {0,1} \right\}^{k \times n}},S^{T}\emph{1}  = \emph{1}
\end{array}
\end{equation}
where $U$ and $S$ are the cluster center matrix and cluster indicator matrix as in (\ref{eq_2}). A difference between the self-paced kmeans and the conventional kmeans is that the cluster centers $U$ is fixed in self-paced approach. This operation can avoid the trivial solution: all samples are trained into a same point in the latent space. $r = \left[ {{r_1}, \ldots ,{r_n}} \right] \in {R^n}$ is a weight vector, $\lambda $ is an age parameter.

Generally, parameter $\lambda $ needs to increase gradually such that more samples can be selected for network training. However, it is difficult to control its growth rate for different tasks. In our work, we adopt a statistic based adaptive approach following~\cite{guo2019adaptive} to update parameter $\lambda $:
\begin{equation}\label{eq_9}
\setlength{\abovedisplayskip}{3pt}
\setlength{\belowdisplayskip}{3pt}
\lambda  = \mu \left( {Klos{s^t}} \right) + {{t\sigma \left( {Klos{s^t}} \right)} \mathord{\left/
 {\vphantom {{t\sigma \left( {Klos{s^t}} \right)} T}} \right.
 \kern-\nulldelimiterspace} T}
\end{equation}
where $Klos{s^t}$ is a loss vector at the $t$-th training step and is calculated as: $Kloss_i^t = \left\| {h_i^{*t} - {U^t}S_{:,i}^t} \right\|_2^2$ for the $i$-th sample. $\mu \left( {Klos{s^t}} \right)$ and $\sigma \left( {Klos{s^t}} \right)$ denote the average and standard deviation of vector $Klos{s^t}$, respectively. $T$ is the maximum training iterations.

By introducing the weight vector $r$ and parameter $\lambda $, the proposed CDIMC-net can select the most confident samples whose clustering losses are no more than $\lambda $ for training. Then with the iteration increases, more confident samples will be selected. This process is similar to the human cognitive learning, \emph{i.e.}, learns from easy to hard or less to more.

\textbf{Loss function of fine-tuning}: Combining the losses of kmeans and graph embedding, the overall loss function of the fine-tuning network is:
\begin{equation}\label{eq_10}
\setlength{\abovedisplayskip}{3pt}
\setlength{\belowdisplayskip}{3pt}
\begin{array}{l}
\mathop {\min }\limits_{S,v,{\Omega _1},\lambda } \frac{1}{nk}\sum\limits_{i = 1}^n {\left( {{r_i}\left\| {h_i^* - U{S_{:,i}}} \right\|_2^2 - \lambda {r_i}} \right)} \\
 + \alpha \frac{1}{{2nl}}\sum\limits_{v = 1}^l {\sum\limits_{i = 1}^n {\sum\limits_{j = 1}^n {\left\| {h_i^{\left( v \right)} - h_j^{\left( v \right)}} \right\|_2^2N_{i,j}^{\left( v \right)}} } } \\
s.t.{\kern 1pt} {\kern 1pt} {r_i} \in \left\{ {0,1} \right\},S \in {\left\{ {0,1} \right\}^{k \times n}},S^{T}\emph{1}  = \emph{1}
\end{array}
\end{equation}

At the end of fine-tuning, CDIMC-net will produce the clustering result $S$ for the incomplete multi-view data.

\subsection{Optimization}
Similar to conventional Autoencoer, all parameters of the pre-training network can be directly optimized by the Stochastic Gradient Descent (SGD) algorithm and back-propagation. Thus, in this section, we focus on the optimization of the fine-tuning network, where an alternating optimization algorithm is adopted to optimize loss function (\ref{eq_10}).

\textbf{Step 1: Update the encoder parameters ${\Omega _1}$}: The optimization problem for encoder parameters ${\Omega _1}$ is:
\begin{equation}\label{eq_11}\small
\begin{array}{l}
\mathop {\min }\limits_{{\Omega _1}} \frac{1}{nk}\sum\limits_{i = 1}^n {\left( {{v_i}\left\| {\frac{{\sum\limits_{v = 1}^l {f_{EC}^{\left( v \right)}\left( {{x_i^{\left( v \right)}}} \right)W_{i,i}^{\left( v \right)}} }}{{\sum\limits_{v = 1}^l {W_{i,i}^{\left( v \right)}} }} - U{S_{:,i}}} \right\|_2^2} \right)} \\
 + \alpha \frac{1}{{2nl}}\sum\limits_{v = 1}^l {\sum\limits_{i = 1}^n {\sum\limits_{j = 1}^n {\left\| {f_{EC}^{\left( v \right)}\left( {x_i^{\left( v \right)}} \right) - f_{EC}^{\left( v \right)}\left( {x_j^{\left( v \right)}} \right)} \right\|_2^2N_{i,j}^{\left( v \right)}} } }
\end{array}
\end{equation}

Problem (\ref{eq_11}) can be adaptively optimized via SGD and back-propagation.

\textbf{Step 2: Update cluster indicator $S$}: $S$ is updated by solving the following problem:
\begin{equation}\label{eq_12}
\setlength{\abovedisplayskip}{3pt}
\setlength{\belowdisplayskip}{3pt}
\mathop {\min }\limits_{S \in {{\left\{ {0,1} \right\}}^{k \times n}},S^{T}\emph{1}  = \emph{1}} {\left\| {H^* - US} \right\|_F^2}
\end{equation}

The optimal solution to problem (\ref{eq_12}) is:
\begin{equation}\label{eq_13}
{S_{i,j}} = \left\{ {\begin{array}{*{20}{c}}
{1,}&{if{\kern 1pt} {\kern 1pt} j = \arg \mathop {\min }\limits_c \left\| {h_i^* - {U_{:,c}}} \right\|_2^2}\\
{0,}&{otherwise}
\end{array}} \right.
\end{equation}

\textbf{Step 3: Update $r$}: The optimization problem to variable $r$ is degraded as follows by fixing the other variables:
\begin{equation}\label{eq_14}
\setlength{\abovedisplayskip}{3pt}
\setlength{\belowdisplayskip}{3pt}
\mathop {\min }\limits_{{r_i} \in \left\{ {0,1} \right\}} \sum\limits_{i = 1}^n {\left( {{r_i}\left\| {h_i^* - U{S_{:,i}}} \right\|_2^2 - \lambda {r_i}} \right)}
\end{equation}

Supposing $Klos{s_i} = \left\| {h_i^* - U{S_{:,i}}} \right\|_2^2$, the optimal solution to problem (\ref{eq_14}) can be expressed as follows:
\begin{equation}\label{eq_15}
{r_i} = \left\{ {\begin{array}{*{20}{c}}
{1,}&{if{\kern 1pt} {\kern 1pt} Klos{s_i} \le \lambda }\\
{0,}&{otherwise}
\end{array}} \right.
\end{equation}

\textbf{Step 4: Update $\lambda $}: $\lambda $ is updated via (\ref{eq_9}).

By alternatively updating the above variables, the proposed CDIMC-net can converge to the local optimal solution.
\subsection{Implementation for IMC}
\begin{algorithm}
  \caption{Fine-tuning and clustering of CDIMC-net}
  \KwIn{Arranged incomplete multi-view data $\left\{ {{Y^{\left( v \right)}}} \right\}_{v = 1}^l$, indicator matrix $\left\{ {{{\bar W}^{\left( v \right)}}} \right\}_{v = 1}^l$, and graphs $\left\{ {{N^{\left( v \right)}}} \right\}_{v = 1}^l$; parameter $\alpha $; Maximum iterations: $T$; Maximum iterations for inner loop: $Maxiter$; Batch size: $b_s$; Stopping threshold: $\xi $.}
  \KwOut{Clustering indicator $S$.}
  \textbf{Initialization:}  Feed $\left\{ {{Y^{\left( v \right)}}} \right\}_{v = 1}^l$, $\left\{ {{{\bar W}^{\left( v \right)}}} \right\}_{v = 1}^l$, and $\left\{ {{N^{\left( v \right)}}} \right\}_{v = 1}^l$ into the pre-trained network to obtain the consensus representation ${H^*}$, and then implement kmeans on it to obtain the initialized cluster center matrix $U$ and cluster indicator matrix $S$. Set all elements of $r$ as 1.\\
  \For{$t \in \left\{ {1,2, \ldots ,T} \right\}$}
  {
    \For{$j \in \left\{ {1,2, \ldots ,Maxiter} \right\}$}
    {
      Update the network parameters by optimizing (\ref{eq_11}) batch to batch;
    }
      Update cluster indicator $S$ using (\ref{eq_13});\\
      Update weight vector $r$ using (\ref{eq_15});\\
      Update step-parameter $\lambda $ using (\ref{eq_9});\\
      \If{$1 - \frac{1}{n}\sum\limits_{i,j} {S_{i,j}^tS_{i,j}^{t - 1}}  < \xi $}
      {
        Stop training\;
      }
    }
  return $S$.\
\end{algorithm}
It should be noted that deep learning commonly groups the given data into several subsets and then feeds these subsets batch by batch for model training. However, for our method, it is difficult to sufficiently utilize the local geometric information of graphs ${N^{\left( v \right)}}$ via the conventional batch-to-batch training approach. To solve this issue, we propose a simple approach to explore such local information as much as possible. Based on the assumption that samples from the same cluster are more likely to have connections marked by edge value `1', we propose to reorder the given samples {\small{$\left\{ {{X^{\left( v \right)}}} \right\}_{v = 1}^l$}} first according to the initialized clustering result obtained by performing kmeans on the features stacked by all views and then construct the nearest neighbor graphs {\small{$\left\{ {{N^{\left( v \right)}}} \right\}_{v = 1}^l$}} from the reordered data, followed by feeding the batch of samples one by one for network training. In this way, every sub-block {\small{$\left\{ {N_{batch}^{\left( v \right)}} \right\}_{v = 1}^l$}} corresponding to the selected batch of samples {\small{$\left\{ {X_{batch}^{\left( v \right)}} \right\}_{v = 1}^l$}} will carry dense nearest neighbor information as much as possible such that more local information can be utilized. Specifically, the detail implementation steps of our CDIMC-net for IMC are presented as follows:

\textbf{Step 1: Data rearrangement and nearest neighbor graph construction}: 1) Concatenating all views into one single view, where the missing instances denoted by `NaN' are filled in the average instance of the corresponding view; 2) Performing kmeans on the stacked view; 3) Reorder data according to the clustering result, where samples grouped into the same cluster are placed together; 4) Construct the nearest neighbor graph according to (\ref{eq_6}) from the reordered data. The rearranged data is denoted by {\small{$\left\{ {{Y^{\left( v \right)}}} \right\}_{v = 1}^l$}}, the corresponding nearest neighbor graph and view indicator matrix are denoted by {\small{$\left\{ {{N^{\left( v \right)}}} \right\}_{v = 1}^l$}} and {\small{$\left\{ {{{\bar W}^{\left( v \right)}}} \right\}_{v = 1}^l$}}, respectively.

\textbf{Step 2: Network pre-training}: Exploit the rearranged data, graphs, and indicator matrices to train the IMC Autoencoder network, where all features of the missing views are set as 0. For each batch, we exploit SGD to optimize the loss function {\small{$\mathop {\min }\limits_{{\Omega _1},{\Omega _2}} \sum\limits_{v = 1}^l {\frac{1}{{{m_v}{b_s}}}\left\| {\left( {Y_{batch}^{\left( v \right)} - \bar Y_{batch}^{\left( v \right)}} \right){{\bar W_{batch}}^{\left( v \right)}}} \right\|_F^2}  + \alpha \frac{1}{{{b_s}l}}\sum\limits_{v = 1}^l {Tr\left( {H_{batch}^{\left( v \right)}{L_{N_{batch}^{\left( v \right)}}}H_{batch}^{\left( v \right)T}} \right)}$}}, where $Y_{batch}^{\left( v \right)}$ denotes the selected batch of data, $\bar Y_{batch}^{\left( v \right)}$ and $H_{batch}^{\left( v \right)}$ are the corresponding reconstructed data and feature representation, ${\bar W_{batch}^{\left( v \right)}}$ and ${L_{N_{batch}^{\left( v \right)}}}$ are the sub-indicator matrix and Laplacian matrix of the $v$-th graph corresponding to the batch of data, ${b_s}$ denotes the batch size.

\textbf{Step 3: Network fine-tuning}: The detailed fine-tuning steps are summarized in Algorithm 1.

\section{Experiment}
\subsection{Experimental Settings}

\textbf{Databases}: Three databases listed in Table 1 are adopted. 1) \textbf{Handwritten}~\cite{asuncion2007uci}: It contains five views and 2000 samples from ten numerals (\emph{i.e.}, 0-9), where the five views are obtained by Fourier coefficients, profile correlations, Karhunen-Love coefficient, Zernike moments, and pixel average extractors. 2) Berkeley Drosophila Genome Project gene expression pattern database (\textbf{BDGP}): BDGP is composed of 5 categories and 2500 samples, where each class has 500 samples~\cite{cai2012joint}. Each sample is represented by four views, \emph{i.e.}, texture feature and three kinds of visual features extracted from the lateral, dorsal, and ventral images. 3) \textbf{MNIST}~\cite{lecun1998mnist}. Following~\cite{wang2018partial}, we evaluate CDIMC-net on the same subset of the MNIST database, which is composed of 4000 samples and ten digits. Pixel feature and edge feature are extracted as two views.
\begin{table}[!t]
\label{tab:1}
\scriptsize
\centering
\begin{tabular}{|c|c|c|c|c|}
\hline
Database	& \# Class	&\# View	&\# Samples	&\# Features\\
\hline
\hline
Handwritten	    &10	&5	&2000	&76/216/64/240/47\\
BDGP        	&5	&4	&2500	&79/1000/500/250\\
MNIST       	&10	&2	&4000	&784/784\\
\hline
\end{tabular}
\caption{Description of the multi-view databases}
\end{table}
\begin{table*}[t!]
\small
  \centering
\resizebox{0.88\textwidth}{!}{
    \begin{tabular}{|c|l|cccc|cccc|}
    \hline
    \multicolumn{2}{|c|}{} & \multicolumn{4}{c|}{ACC}      & \multicolumn{4}{c|}{NMI} \\
    \hline
    \multicolumn{1}{|c|}{Database} & Method\textbackslash{}$p\%$ & 0.1   & 0.3   & 0.5   & 0.7   & 0.1   & 0.3   & 0.5   & 0.7 \\
    \hline
    \multirow{8}[2]{*}{\begin{turn}{-270}Handwritten\end{turn}}
&BSV	&68.27$\pm$5.66	&51.49$\pm$2.29	&38.24$\pm$2.25	&27.15$\pm$1.31	&62.82$\pm$3.24	&47.01$\pm$1.71	&32.21$\pm$1.00	&19.48$\pm$0.69\\
&Concat	&75.06$\pm$3.86	&55.48$\pm$1.57	&42.19$\pm$0.99	&28.31$\pm$0.75	&73.08$\pm$2.05	&51.66$\pm$0.99	&38.24$\pm$1.59	&23.50$\pm$0.95\\
&MIC	&77.59$\pm$2.41	&73.29$\pm$3.41	&61.27$\pm$3.16	&41.34$\pm$2.69	&70.84$\pm$2.08	&65.39$\pm$2.08	&52.95$\pm$1.33	&34.71$\pm$2.11\\
&OMVC	&65.04$\pm$6.50	&55.00$\pm$5.06	&36.40$\pm$4.93	&29.80$\pm$4.63	&56.72$\pm$5.05	&44.99$\pm$4.56	&35.16$\pm$4.62	&25.83$\pm$8.37\\
&DAIMC	&88.86$\pm$0.63	&86.73$\pm$0.79	&81.92$\pm$0.88	&60.44$\pm$6.87	&79.78$\pm$0.71	&76.65$\pm$1.07	&68.77$\pm$0.99	&47.10$\pm$4.79\\
&OPIMC	&80.20$\pm$5.40	&76.45$\pm$5.15	&69.50$\pm$6.54	&56.66$\pm$10.06	&77.26$\pm$3.11	&73.74$\pm$3.42	&66.57$\pm$4.18	&51.86$\pm$7.97\\
&MKKM-IK-MKC	&71.78$\pm$1.74	&69.07$\pm$0.73	&66.08$\pm$3.25	&55.55$\pm$1.39	&69.43$\pm$1.28	&65.42$\pm$0.61	&59.04$\pm$2.69	&47.36$\pm$1.78\\
&CDIMC-net	&\textbf{95.12$\pm$1.11}	&\textbf{94.22$\pm$1.07}	&\textbf{91.48$\pm$0.82}	&\textbf{88.85$\pm$0.77}	&\textbf{90.10$\pm$1.97}	&\textbf{89.21$\pm$1.11}	&\textbf{84.58$\pm$1.08}	&\textbf{79.99$\pm$0.78}\\
    \hline
    \multirow{8}[2]{*}{\begin{turn}{-270}BDGP\end{turn}}
&BSV	&51.48$\pm$3.96	&41.44$\pm$3.55	&34.74$\pm$1.52	&27.95$\pm$1.76	&35.74$\pm$4.01	&25.20$\pm$2.70	&16.39$\pm$1.36	&9.29$\pm$1.67\\
&Concat	&57.66$\pm$4.79	&50.04$\pm$1.58	&40.41$\pm$3.52	&27.52$\pm$1.24	&44.58$\pm$4.78	&31.81$\pm$1.45	&19.76$\pm$1.78	&6.17$\pm$1.27\\
&MIC	&48.31$\pm$0.83	&40.88$\pm$1.18	&34.02$\pm$1.42	&29.45$\pm$0.91	&28.52$\pm$0.49	&23.94$\pm$1.21	&11.05$\pm$0.89	&7.04$\pm$1.17\\
&OMVC	&55.23$\pm$4.55	&46.22$\pm$3.15	&39.46$\pm$1.12	&38.32$\pm$2.95	&28.78$\pm$1.59	&19.44$\pm$1.20	&13.51$\pm$1.21	&12.74$\pm$5.46\\
&DAIMC	&77.34$\pm$2.58	&69.30$\pm$6.42	&52.45$\pm$8.57	&35.68$\pm$4.23	&55.64$\pm$2.68	&47.87$\pm$4.65	&28.33$\pm$1.38	&9.17$\pm$3.64\\
&OPIMC	&79.38$\pm$7.69	&63.73$\pm$7.29	&55.17$\pm$9.24	&35.82$\pm$2.68	&61.77$\pm$7.41	&41.47$\pm$3.88	&25.94$\pm$7.29	&8.35$\pm$2.23\\
&MKKM-IK-MKC	&31.80$\pm$1.68	&29.12$\pm$0.29	&29.44$\pm$1.39	&35.16$\pm$1.69	&7.21$\pm$1.10	&5.86$\pm$0.56	&6.65$\pm$1.19	&12.19$\pm$1.41\\
&CDIMC-net	&\textbf{89.02$\pm$0.71}	&\textbf{77.99$\pm$1.00}	&\textbf{62.14$\pm$1.67}	&\textbf{40.98$\pm$1.05}	&\textbf{77.24$\pm$1.98}	&\textbf{57.09$\pm$0.77}	&\textbf{35.64$\pm$0.59}	&\textbf{14.77$\pm$0.99}\\
    \hline
    \multirow{12}[2]{*}{\begin{turn}{-270}MNIST\end{turn}}
&BSV	        &33.25$\pm$1.79	&37.37$\pm$0.69	&42.76$\pm$1.30	&47.95$\pm$1.36	&27.20$\pm$0.89	&31.39$\pm$1.28	&37.45$\pm$1.42	&42.49$\pm$1.47\\
&Concat	    &36.88$\pm$1.96	&39.24$\pm$1.47	&43.79$\pm$1.71	&47.37$\pm$1.08	&34.48$\pm$1.09	&33.38$\pm$0.54	&37.42$\pm$1.38 &43.17$\pm$0.69	\\
&PMVC	    &41.36$\pm$2.29	&43.42$\pm$2.99	&44.68$\pm$1.23	&45.84$\pm$1.59	&35.46$\pm$0.25	&38.51$\pm$1.63	&39.43$\pm$1.37	&39.83$\pm$1.71	\\
&IMG		    &46.34$\pm$3.36 &47.13$\pm$2.24	&46.88$\pm$1.51	&48.31$\pm$1.22	&39.74$\pm$2.42 &40.71$\pm$2.56	&39.87$\pm$1.05	&44.16$\pm$1.09\\
&IMC\_GRMF	&49.12$\pm$2.46	&50.59$\pm$2.59	&52.37$\pm$1.67	&52.46$\pm$1.59	&47.36$\pm$0.97	&48.18$\pm$1.57	&50.73$\pm$0.98	&51.57$\pm$1.03	\\
&MIC	        &43.96$\pm$2.38	&44.42$\pm$2.28	&44.17$\pm$1.37	&45.38$\pm$2.82	&38.77$\pm$1.35	&40.81$\pm$1.28	&40.53$\pm$0.67	&41.61$\pm$1.50	\\
&OMVC	    &40.44$\pm$2.95	&42.23$\pm$2.17	&40.36$\pm$2.20	&41.44$\pm$3.39	&36.21$\pm$1.47	&36.68$\pm$2.16	&35.64$\pm$1.89	&32.25$\pm$2.95	\\
&DAIMC	    &45.33$\pm$4.12	&48.19$\pm$1.38	&49.25$\pm$1.67	&49.36$\pm$1.87	&37.46$\pm$3.04	&41.09$\pm$1.58	&43.47$\pm$0.82	&44.15$\pm$0.75	\\
&OPIMC	    &41.40$\pm$2.51	&48.02$\pm$2.63	&47.77$\pm$3.39	&48.71$\pm$2.44	&34.29$\pm$2.33	&43.98$\pm$1.98	&44.63$\pm$1.47	&45.65$\pm$1.15	\\
&MKKM-IK-MKC	&47.56$\pm$2.18	&51.02$\pm$0.66	&51.72$\pm$0.58	&52.45$\pm$0.41	&40.39$\pm$1.17	&42.76$\pm$0.70	&43.88$\pm$0.54	&45.10$\pm$0.39	\\
&PMVC\_CGAN	&45.17$\pm$-\,-	&48.36$\pm$-\,-	&52.80$\pm$-\,-	&52.02$\pm$-\,-	&39.33$\pm$-\,-	&43.22$\pm$-\,-	&49.61$\pm$-\,-	&48.22$\pm$-\,- \\
&CDIMC-net	&\textbf{51.65$\pm$0.14} &\textbf{57.64$\pm$1.44} &\textbf{58.28$\pm$0.68} &\textbf{59.15$\pm$0.21} &\textbf{48.25$\pm$0.47} &\textbf{50.54$\pm$1.26} &\textbf{51.70$\pm$0.67} &\textbf{52.87$\pm$0.48}\\
    \hline
    \end{tabular}}
\caption{Clustering average results and standard deviations of different methods on the Handwritten, BDGP, and MNIST databases with different missing-view rates or paired-view rates $p\%$. Note: the average results of PMVC\_CGAN are reported in [Wang \emph{et al}., 2018].}
  \label{tab:addlabel}
\end{table*}

\textbf{Compared methods}: Compared methods include: PMVC, IMG, MIC, OMVC, DAIMC, OPIMC, IMC with graph regularized matrix factorization (IMC\_GRMF)~\cite{wen2018incomplete}, MKKM-IK-MKC, and PMVC\_CGAN. Besides, two baseline methods, \emph{i.e.}, best single view (BSV)~\cite{zhao2016incomplete} and Concat~\cite{zhao2016incomplete} are also evaluated, where BSV reports the results of the best view, and Concat implements the kmeans on the stacked features of all views. For CDIMC-net, the encoder and decoder networks are stacked by four full connected layers with size of $[0.8m_v,0.8m_v,1500,k]$ and $[k,1500,0.8m_v,0.8m_v]$, respectively. The activation function is `ReLU' and the optimizer is `SGD' for the pre-training network and `ADAM' for the fine-tuning network. CDIMC-net is implemented on PyTorch and Ubuntu Linux 16.04.

\textbf{Incomplete data construction}: For the data with more than two views, we randomly remove $p\%$ ($p\in \{10,30,50,70\}$) instances from every view under the condition that all samples at least have one view. For MNIST database, $p\%$ ($p\in \{10,30,50,70\}$) instances are randomly selected as paired samples whose views are complete, and the remaining samples are treated as single view samples, where half of them only have the first view and the other half of the samples only have the second view.

\textbf{Evaluation metric}: Clustering accuracy (ACC) and normalized mutual information (NMI)~\cite{hu2018doubly}.
\subsection{Experimental Results and Analysis}
Experimental results on the above three databases are listed in Tables 2. We can observe the following points from the results: 1) CDIMC-net significantly outperforms the other methods on the three databases. For instance, on the Handwritten database with a missing-view rate of 50\%, CDIMC-net obtains about 91\% ACC and 84\% NMI, which are about 10\% and 16\% higher than those of the second best method, respectively. 2) BSV and Concat obtain worse IMC performance than the other methods in most cases. Thus, we can conclude that exploring the complementary information and consistent information of multiple views is beneficial to improve the performance. 3) CDIMC-net performs better than the advanced deep network based method PMVC\_CGAN on the MNIST database. This demonstrates that CDIMC-net is superior to PMVC\_CGAN for IMC.

\subsection{Parameter Analysis}
Fig.2 shows the relationships of ACC, graph embedding hyper-parameter $\alpha$, and the learning rate of the CDIMC-net on the Handwritten and BDGP databases with a missing-view rate of 10\%. We can observe that CDIMC-net obtains relatively better performance when the two parameters are small. Specifically, a large learning rate is harmful to obtain the local optimal clustering results and a large $\alpha$ makes the graph embedding term dominate the training phase. In the applications, we suggest selecting the learning rate and $\alpha$ from [1e-5,1e-3].

\subsection{Component Analysis}
In this subsection, we conduct experiments on the Handwritten and BDGP databases to validate the importance of graph embedding, self-paced learning, and pre-training, where the degenerate models of CDIMC-net without a graph embedding term, self-pace constraint, and pre-training phase, are compared. From Fig.3, we can find that CDIMC-net outperforms the three degenerate models, which demonstrates the effectiveness of the introduced three approaches.
\begin{figure}[!htb]
\centering
\subfloat[Handwritten]{\includegraphics[width=1.4in,height=0.9in]{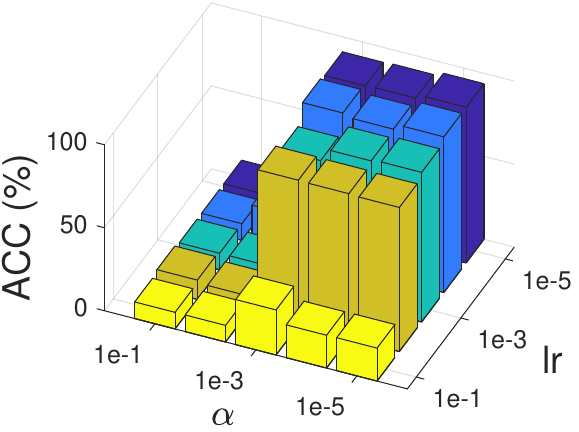}}
\hfil
\subfloat[BDGP]{\includegraphics[width=1.4in,height=0.9in]{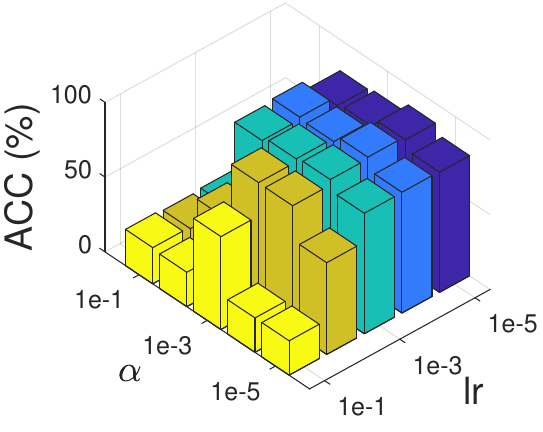}}
\caption{ACC (\%) v.s. $\alpha$ and learning rate (lr) of CDIMC-net on (a) Handwritten and (b) BDGP databases with a missing-view rate of 10\%.}
\label{fig:fig3}
\end{figure}
\begin{figure}[!htb]
\centering
\subfloat[Handwritten]{\includegraphics[width=1.5in,height=1in]{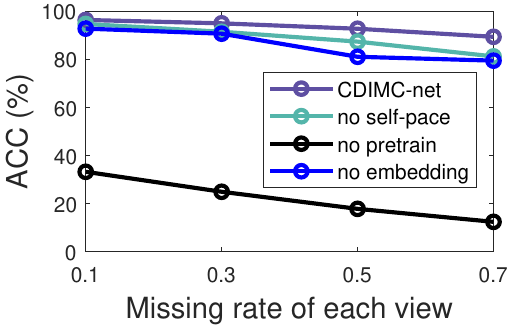}}
\hfil
\subfloat[BDGP]{\includegraphics[width=1.5in,height=1in]{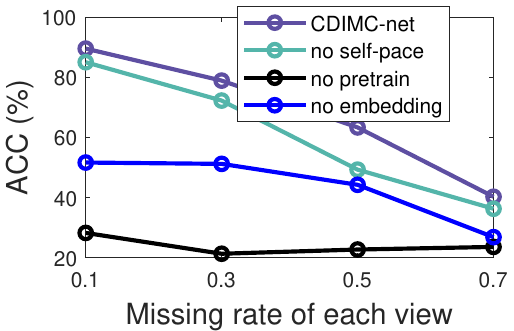}}
\caption{ACC (\%) of CDIMC-net and its three degenerate models on (a) Handwritten and (b) BDGP databases.}
\label{fig:fig4}
\end{figure}
\begin{figure}[!htb]
\centering
\subfloat[Handwritten]{\includegraphics[width=1.4in,height=0.9in]{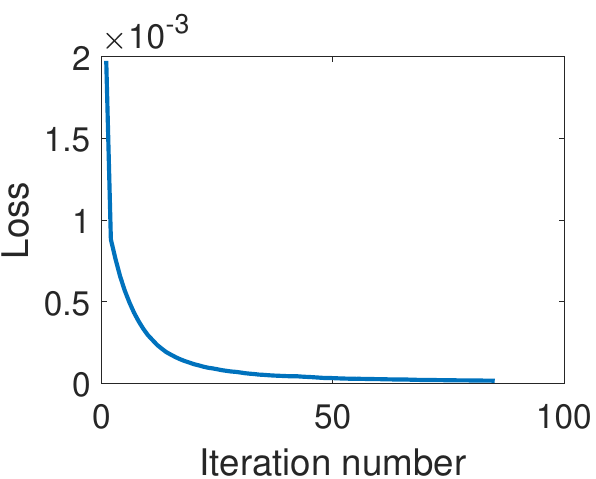}}
\hfil
\subfloat[BDGP]{\includegraphics[width=1.4in,height=0.8in]{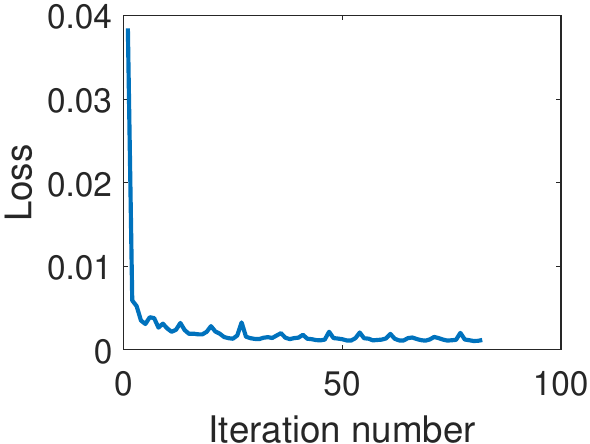}}
\caption{Loss v.s. iterations of CDIMC-net on (a) Handwritten and (b) BDGP databases with a missing-view rate of 10\%.}
\label{fig:fig4}
\end{figure}
\subsection{Convergence Analysis}
Fig.4 shows the loss value of fine-tuning network versus the iterations on the Handwritten and BDGP databases with a missing-view rate of 10\%. From the figures, we can observe that the loss value shows a downward trend overall and decreases quickly in the first few steps. This validates the convergence property of the proposed method.
\section{Conclusion}
In this paper, we proposed a novel and flexible CDIMC-net, which can handle all kinds of incomplete data. Based on human cognitive learning, CDIMC-net introduces the self-paced kmeans to improve the robustness to outliers. Besides this, it incorporates the graph embedding technique to preserve the local structure of data. The superior performance of CDIMC-net are validated on several incomplete cases with the comparison of many state-of-the-art IMC methods.
\section*{Acknowledgements}
This work is partially supported by Shenzhen Fundamental Research Fund under Grant no. JCYJ20190806142416685, Guangdong Basic and Applied Basic Research Foundation under Grant nos. 2019A1515110582 \& 2019A1515110475, Establishment of Key Laboratory of Shenzhen Science and Technology Innovation Committee under Grant no. ZDSYS20190902093015527, University of Macau (File no. MYRG2019-00006-FST), National Natural Science Foundation of China under Grant nos. 61702110 \& 61702163, and National Postdoctoral Program for Innovative Talent under Grant no. BX20190100.

\bibliographystyle{named}
\bibliography{ijcai20}

\end{document}